\documentclass{article}

\usepackage{microtype}
\usepackage{graphicx}
\usepackage{subcaption}
\usepackage{booktabs} 

\usepackage{hyperref}

\usepackage[accepted]{icml2026}

\usepackage{amsmath}
\usepackage{amssymb}
\usepackage{mathtools}
\usepackage{amsthm}
\usepackage{multirow}
\usepackage{makecell}
\usepackage{placeins}

\usepackage{dblfloatfix}
\usepackage{afterpage}

\usepackage[capitalize,noabbrev]{cleveref}

\theoremstyle{plain}

\theoremstyle{definition}

\theoremstyle{remark}

\usepackage[textsize=tiny]{todonotes}

\icmltitlerunning{Submission and Formatting Instructions for ICML 2026}

\begin{document}

\twocolumn[
  \icmltitle{Affine-Scaled Attention: Towards Flexible and Stable Transformer Attention }  



  \icmlsetsymbol{equal}{*}

  \begin{icmlauthorlist}
    \icmlauthor{Jeongin Bae}{equal,naver}
    \icmlauthor{Baeseong Park}{equal,naver}
    \icmlauthor{Gunho Park}{naver}
    \icmlauthor{Minsub Kim}{naver}
    \icmlauthor{Joonhyung Lee}{naver}
    \icmlauthor{Junhee Yoo}{naver}
    \icmlauthor{Sunghyeon Woo}{naver}
    \icmlauthor{Jiwon Ryu}{naver}
    \icmlauthor{Se Jung Kwon}{naver}
    \icmlauthor{Dongsoo Lee}{naver}
  \end{icmlauthorlist}

  \vspace{2mm}
  {\centering
    {\normalsize NAVER Cloud\par}
    \vspace{1pt}
    {\normalsize \textnormal{\{jeong\_in.bae, baeseong.park\}@navercorp.com}\par}
  }
  \vspace{2.0mm}



  \icmlkeywords{Machine Learning, ICML}

   \addvspace{1.5mm}
]



\printAffiliationsAndNotice{\icmlEqualContribution}

\begin{abstract}
  Transformer attention is typically implemented using softmax normalization, which enforces attention weights with unit sum normalization. While effective in many settings, this constraint can limit flexibility in controlling attention magnitudes and may contribute to overly concentrated or unstable attention patterns during training. Prior work has explored modifications such as attention sinks or gating mechanisms, but these approaches provide only limited or indirect control over attention reweighting. We propose Affine-Scaled Attention, a simple extension to standard attention that introduces input-dependent scaling and a corresponding bias term applied to softmax-normalized attention weights. This design relaxes the strict normalization constraint while maintaining aggregation of value representations, allowing the model to adjust both the relative distribution and the scale of attention in a controlled manner. 
 
  We empirically evaluate Affine-Scaled Attention in large-scale language model pretraining across multiple model sizes. Experimental results show consistent improvements in training stability, optimization behavior, and downstream task performance compared to standard softmax attention and attention sink baselines. These findings suggest that modest reweighting of attention outputs provides a practical and effective way to improve attention behavior in Transformer models.
\end{abstract}

\section{Introduction}

Transformer-based models have become a foundational architecture for sequence modeling~\cite{wang2024understandingexpressivepowermechanisms, jiang2025approximationratetransformerarchitecture}, achieving state-of-the-art performance across a wide range of natural language processing tasks~\cite{islam2023comprehensivesurveyapplicationstransformers, minaee2025largelanguagemodelssurvey}. 
A central component of these models is the softmax attention mechanism~\cite{vaswani2017attention}, which converts dot-product attention scores to attention weights by normalizing them with a softmax function.
Such a normalization process enables a weighted aggregation of contextual information, allowing the model to capture dependencies among tokens effectively~\cite{bahdanau2016neuralmachinetranslationjointly, luong2015effectiveapproachesattentionbasedneural}.
The empirical and theoretical effectiveness of softmax attention has been demonstrated in large-scale language models~\cite{qin2022cosformerrethinkingsoftmaxattention, choromanski2022rethinkingattentionperformers, collins2024context, deng2023superiority, mongaras2025expressiveness, han2024bridging}.

Despite its success, recent works have highlighted the limitations of the standard softmax formulation~\cite{richter2020normalizedattentionprobabilitycage, Miller2023AttentionOffByOne, xiao2023efficient}. 
In particular, softmax inherently induces competition among tokens. Since the attention weights are normalized, the model allocates probability mass across contextual tokens even when no tokens provide meaningful information~\cite{Miller2023AttentionOffByOne}.
This suggests that the normalization constraint may be inappropriate when the model should \emph{attend to nothing} in certain contexts.
Moreover, prior studies report that attention often assigns large fraction of attention weights to early tokens regardless of semantic relevance, a behavior attributed in part to softmax normalization enforcing a fixed total mass across positions~\cite{xiao2023efficient, gu2025attentionsinkemergeslanguage, barbero2025llmsattendtoken, ruscio2025sinkinggeometricapproachattention}.



To mitigate these effects, several attention modifications have been proposed. 
\citet{Miller2023AttentionOffByOne} adds a constant term to the softmax denominator, allowing the attention distribution to concentrate on an additional \emph{null} mass rather than only among input tokens. 
This modification is referred to as \emph{off-by-one} attention, since it effectively introduces an extra sink term of one into the normalization.
Building on a closely related intuition, GPT-OSS~\cite{agarwal2025gpt} incorporates an attention-sink mechanism in which each attention head learns a bias term that effectively introduces a learnable \emph{sink} in the softmax normalization. 
This design enables the model to assign negligible attention to all contextual tokens when doing so is appropriate. 


Gated Attention~\cite{qiu2025gated} takes a different approach by introducing a widely used gating mechanism to improve attention computation.
Without modifying the softmax normalization, it augments the scaled dot-product attention (SDPA) output with an explicit gating mechanism. The gate modulates how much of the attended context is propagated forward. 
Empirically, this gating reduces the fraction of attention effectively devoted to the first token.

These prior approaches, while effective in some respects, leave key limitations unresolved. 
In off-by-one and attention sink designs, the additional sink term is typically independent of the input hidden states and is treated as either a constant or a learnable parameter.
This raises the question of whether such an input-agnostic sink can reliably stabilize attention across diverse contexts. 
Meanwhile, Gated Attention gates the post-attention representation and therefore does not remove the unit-sum constraint imposed by softmax. 
Even if it reduces the influence of the first token, the attention distribution is still forced to allocate a fixed total weight over tokens, potentially preserving over-concentration.

Accordingly, these observations motivate a natural question: does the current formulation provide an optimal mechanism for stabilizing attention while retaining sufficient expressiveness? To answer this, we investigate extensions to the softmax operation itself, aiming to improve both the stability and the expressiveness of attention.

In this paper, we propose \textsc{Affine-Scaled Attention}, which augments attention weights with an input-dependent scaling factor and an additional bias term. This formulation relaxes the requirement that softmax-normalized weights sum to one, thereby increasing the expressiveness of attention and improving performance. The bias term further introduces a controllable offset within scaled dot-product attention, mitigating the overly competitive, \emph{winner-takes-all} behavior often induced by softmax. Together, these components yield more flexible and stable attention distributions that better reflect the preferences of the model. 

\begin{figure}[t]
  \vskip 0.2in
  \centering
  
  \includegraphics[width=0.48\textwidth]{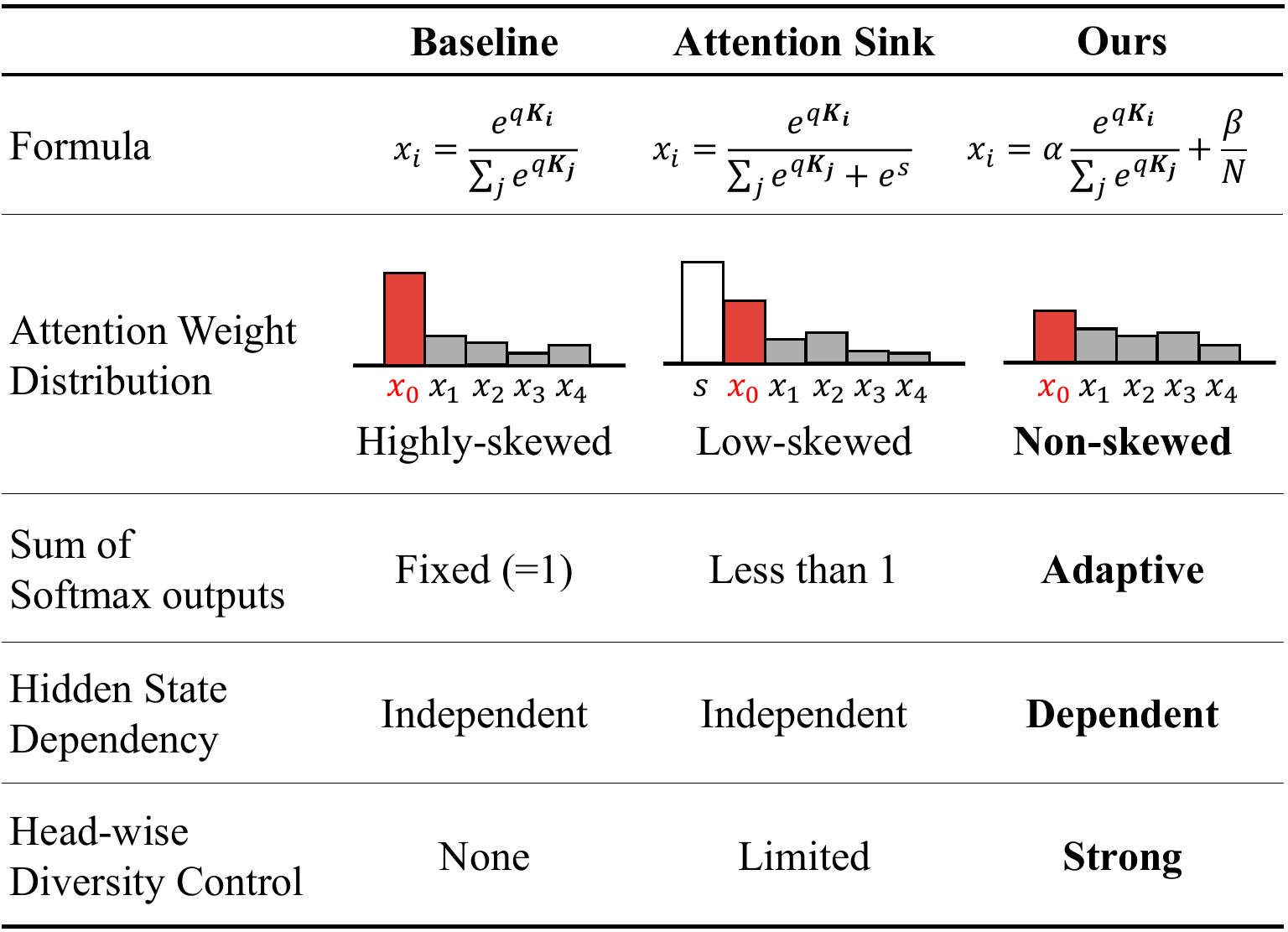}

  \caption{%
    Comparison of baseline softmax, attention sink, and Affine-Scaled Attention. Attention skew toward the first token (red) is progressively reduced. Here, $i$ denotes the index over key tokens, and the attention weight distributions are illustrated for a single query.
  }
  \label{fig:main}
\end{figure}

As shown in Figure~\ref{fig:main}, our method produces a less skewed attention weights than standard softmax, enables adaptive per-query scaling unlike attention sink, and promotes greater head diversity through hidden state dependent modulation.

Our major contributions are as follows:

 \begin{itemize}
    \item Provide an in-depth analysis of softmax behavior in decoder-only language models by tracking the evolution of attention logits.
    \item Introduce Affine-Scaled Attention, which applies a scaling factor and a bias to the softmax-normalized attention weights to relax the unit-sum constraint.
    \item Analyze attention allocation dynamics, showing reduced first-token bias, more diverse head utilization and increased attention entropy with our method.
\end{itemize}


\section{Background}

\subsection{LLM training}

Several frontier-scale models such as Llama~\cite{grattafiori2024llama3herdmodels}, DeepSeek-V3~\cite{deepseekai2025deepseekv3technicalreport}, and GPT-OSS~\cite{agarwal2025gpt}, have demonstrated strong capabilities in language understanding and generation. Despite this progress, training even moderately sized models (e.g., 1B–3B parameters) from scratch remains computationally expensive and time-consuming. Consequently, efficient learning paradigms that transfer knowledge from large foundation models have emerged as practical alternatives.

Prior works has shown that distilling or transferring knowledge from large-scale pretrained models can enable smaller models to converge faster and achieve competitive performance compared to training from scratch. For example, Distilling Step-by-Step~\cite{hsieh2023distilling} demonstrates that student models with substantially fewer parameters can attain strong performance by distilling the reasoning processes of large LLMs. More broadly, this paradigm allows relatively compact models to exploit the rich representations learned by large foundation models, thereby improving training efficiency under constrained computational budgets. In this study, we also incorporate knowledge distillation to facilitate the training of moderately sized models.

\subsection{Softmax attention}




In Transformer architectures, attention is computed from three sets of representations: queries ($Q$), keys ($K$), and values ($V$). 
Given a query, attention weights are obtained by taking scaled dot products between the query and all keys, dividing by the square root of the key dimension ($\sqrt{d_k}$), and normalizing the resulting scores with the softmax. 
The normalized attention weights are used to compute a weighted sum of the value vectors, producing the attention output:
\begin{equation}
\operatorname{Attn}(Q, K, V)
= \operatorname{softmax}\left( \frac{QK^{\top}}{\sqrt{d_k}} \right) V \, .
\label{eq:scaled-dot-product-attention}
\end{equation}

This formulation can be viewed as selecting how much each value representation contributes to the final output based on query--key interactions. The softmax operation converts these interaction scores into nonnegative weights that sum to one, forming a probability distribution over tokens.

While effective, this normalization imposes structural constraints. Because the attention weights must sum to one, the model cannot uniformly reduce all weights at once, even in contexts where no token should be strongly attended to. In addition, the exponential mapping in softmax can produce highly peaked distributions, which may lead to excessive concentration of attention on a small number of tokens.

\subsection{Attention sink}

The motivation to revisit the softmax function stems from a structural limitation of the standard formulation. Softmax enforces a forced choice among competing alternatives: even when the model should assign negligible weight to all options, the output is still constrained to sum to one. When applied to attention, this simplex constraint can distort the allocation of probability mass across tokens.

To address this issue, \citet{Miller2023AttentionOffByOne} proposed \emph{Off-by-One} attention, which adds a constant term to the softmax denominator and thereby relaxes the normalization constraint. The resulting operator, denoted $\operatorname{softmax}_{one}$, is defined for $x \in \mathbb{R}^d$, $i \in \{1,\ldots,d\}$ as
\begin{equation}
{\operatorname{softmax}_{one}(x)}_i
= \frac{e^{x_i}}{\sum_{j=1}^{d} e^{x_j} + 1} .
\label{eq:softmax-off-by-one}
\end{equation}
By introducing an additive constant in the denominator, $\operatorname{softmax}_{one}$ retains the ability to produce near-standard softmax behavior when appropriate, while also allowing the overall magnitude of the outputs to be uniformly reduced. Importantly, this modification preserves the relative ordering of logits induced by the original softmax.

A related empirical observation is the \emph{attention sink} phenomenon reported in StreamingLLM~\cite{xiao2023efficient}. In the context of sliding-window attention, the authors show that retaining only the key--value (KV) pairs corresponding to a small set of initial tokens can recover much of the performance loss. They attribute this behavior to a tendency in the model to allocate disproportionately large attention mass to first tokens, which act as sinks even when they are not semantically informative. Preserving a small number of such sink tokens stabilizes the attention distribution and mitigates drift from the full-context attention pattern, enabling efficient long-context inference by combining sink-token KV states with those of the most recent tokens.

Motivated by these behaviors, Off-by-One attention has also been incorporated into the training of GPT-OSS~\cite{agarwal2025gpt}, where it is reported to improve training stability. 
GPT-OSS replaces the fixed constant with a learnable scalar sink parameter $s$, which contributes an additional term $e^{s}$ in the denominator. We denote this variant as $\operatorname{softmax}_{sink}$:
\begin{equation}
{\operatorname{softmax}_{sink}(x)}_i
= \frac{e^{x_i}}{\sum_{j=1}^{d} e^{x_j} + e^{s}} \, .
\label{eq:attention sink}
\end{equation}
Throughout this paper, we use the term \emph{attention sink} to refer to this learnable-denominator formulation.

While the attention sink relaxes the simplex constraint of softmax, it provides limited control over the resulting weights. 
The additional sink term is input-independent, which restricts expressiveness and limits the model's ability to adapt the degree of attenuation to the context.

\subsection{Gated Attention}
Another line of work augments attention with an explicit gating mechanism. 
Gated Attention~\cite{qiu2025gated} observes that gating is widely used in neural architectures, yet its role within attention has not been systematically studied. 
The authors analyze variants by changing where the gate is applied and show that a simple design consistently improves performance. In particular, applying a head-specific sigmoid gate after the scaled dot-product attention (SDPA) output yields reliable gains across settings.

Formally, let $X$ denote the hidden states, $W_a$ the gating weights, and $\phi$ an activation function. Gated Attention applies a multiplicative gate to the SDPA output:
\begin{equation}
\operatorname{Attn_{gated}}(Q,K,V,X) =\operatorname{Attn}(Q,K,V)\cdot G(X),\\
\label{eq:gated-attention}
\end{equation}
\begin{equation}
G(X) = \phi(W_a X).
\end{equation}
By introducing a multiplicative gate, this formulation adds nonlinearity and promotes sparsity in the effective attention output. 
The authors also report that the learned gates tend to reduce the influence of the first token.

Conceptually, gated attention acts as a post-attention filter that modulates how much attended information is propagated. Because it does not directly alter the normalized attention weights, it offers less explicit control over fine-grained reweighting at the level of individual token pairs.





\section{Methodology}

In this section, we propose \textsc{Affine-Scaled Attention}, a modified formulation that augments softmax attention with an input-dependent scaling factor and an associated bias term. This design relaxes the unit-sum constraint and mitigates overly competitive, \emph{winner-takes-all} behavior, leading to more stable training and improved performance. We first analyze behaviors induced by existing softmax-based attention formulations that motivate our approach, and then introduce our modified softmax that increases flexibility.


\subsection{Motivation}

\begin{figure}[t]
  \vskip 0.2in
  \centering

  \includegraphics[width=0.45\textwidth]{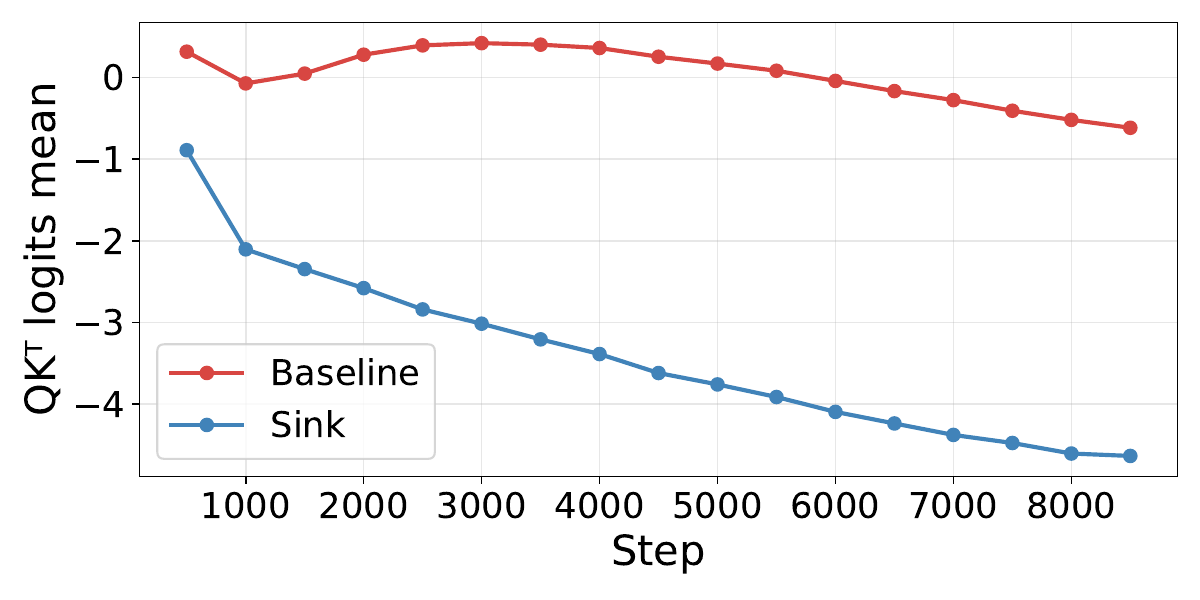}

  \caption{%
    Training dynamics of attention logits. $QK^T$ logits mean over training steps for the 3B baseline and attention sink models, averaged across layers, heads, and token positions.
  }
  \label{fig:qkt_graph}
\end{figure}

To better understand how existing softmax-based attention variants influence training dynamics, we track the evolution of attention logits during pretraining. 
Concretely, we monitor the magnitude of the dot-product term $QK^{\top}$ and compute its mean value aggregated over all layers, heads, and token positions. 
We observe that the average magnitude of $QK^{\top}$ steadily decreases over training (Figure~\ref{fig:qkt_graph}). 

This observation suggests that the scale of query--key alignment becomes smaller over time, implying that pairwise token-to-token interaction scores are progressively attenuated before normalization. 
One plausible interpretation is that the model implicitly learns to reduce the strength of competitive interactions induced by softmax, effectively pushing attention logits toward a lower-magnitude regime. 
The sink term further amplifies this behavior by introducing an additional mass in the softmax denominator. As a result, attention weights are no longer constrained to allocate their entire unit mass solely across tokens via the $\exp(QK^{\top}/\sqrt{d_k})$ terms, which allows the model to decrease individual $QK^{\top}$ values during training.

However, standard softmax attention imposes a structural unit-sum constraint across tokens. Even if the model benefits from globally weaker attention, it cannot uniformly reduce all weights. 
Instead, it must redistribute a fixed unit mass, which can preserve undesirable concentration. 
Importantly, this tendency is not fully resolved by introducing a sink term, since the sink primarily attenuates the overall attention magnitude while leaving the relative allocation among tokens largely unchanged. 
This mismatch between the preference for reduced logit magnitudes and the unit-sum constraint motivates relaxing the normalization itself.

\subsection{Affine-Scaled Attention}
\label{sub:methodology}
 
To relax the constraints imposed by the softmax normalization and to better accommodate the preference for reduced attention magnitudes, we propose Affine-Scaled Attention. This method enhances the attention mechanism with a scaling term and an associated bias term derived from this scaling factor, both of which are applied directly to the attention weights.
By introducing these additional degrees of freedom, Affine-Scaled Attention enables the model to modify the overall magnitude of attention more flexibly, without being strictly constrained by the standard softmax formulation that forces the terms to sum up to one.
Formally, Affine-Scaled Attention is defined as follows:
\begin{equation}
\begin{split}
\operatorname{Attn_{affine}}&(Q,K,V,X) \\ & =\Bigl[\alpha(X)\,\operatorname{softmax}\!\Bigl(\tfrac{QK^\top}{\sqrt{d_k}}\Bigr)
+\beta(X)\Bigr]\,V,
\end{split}
\label{eq:reweighted-attention}
\end{equation}
\begin{equation}
\alpha(X) = \phi(W_a X),
\label{eq:alpha}
\end{equation}
\begin{equation}
\beta(X) = \frac{\alpha_{\text{ma}} - \alpha(X)}{N}.
\label{eq:beta}
\end{equation}
Given hidden state representations $X$, a learnable gating parameter $W_\alpha$ is applied to make each of the scale. The activation function $\phi$ introduces nonlinearity, whereas $\alpha$ and $\beta$ correspond to the scale and bias transformations, respectively. Here, $N$ denotes the key sequence length, which is used to normalize the bias by sequence length.



In Affine-Scaled Attention, we modulate the scaling terms via a gating mechanism to adapt to varying input representations. 
Concretely, gate weights are computed from the hidden states and passed through a nonlinear activation, producing input-dependent scaling parameters. 
Since gating has been shown to be effective across a broad range of neural architectures, we adopt it here to improve the expressiveness and flexibility of our attention formulation by allowing the modulation to adjust dynamically as a function of the input. 
The resulting scaling expands the dynamic range of the attention weights, thereby enabling finer-grained control over the relative contribution of value representations. 
However, an overly expanded range can introduce excessive variability in the attention weights, which may weaken the consistent influence of value information. 
To mitigate this effect, we introduce an additional bias term that regulates the overall scale of the attention weights, ensuring that value representations remain reflected in the attention output.



This bias term is defined as the deviation of the current scaling factor from its moving average. We maintain the moving average as an exponential running estimate:
\begin{equation}
\begin{gathered}
\alpha_{\text{ma},0}=0 \\
\alpha_{\text{ma},t}
=
\rho \alpha_{\text{ma},t-1}
+
(1-\rho)\mathbb{E}\left[\alpha_t(X)\right],
\end{gathered}
\label{eq:center-eq}
\end{equation}
where $\mathbb{E}[\alpha_t(X)]$ denotes the batch-wise mean of the scaling factor and $\rho \in [0,1]$ is a momentum coefficient that controls the trade-off between the running estimate and the current batch statistic.
Equation~\ref{eq:beta} produces a bias that compensates for shifts in the typical magnitude of $\alpha_t(X)$, thereby stabilizing the scale of the attention weights. 
The bias is divided by the key sequence length $N$ to ensure its magnitude remains comparable across different sequence lengths.

Finally, to prevent the scaling factors from collapsing to a specific value and to preserve input-adaptive modulation, we apply a custom nonlinear function, described in Section~\ref{sub:activation_function}.






\section{Attention allocation dynamics}

In this section, we present a comparative analysis of Affine-Scaled Attention and discuss how it leads to performance improvements. Our analysis compares trained models across several attention methods, including the baseline with standard softmax, attention sink, and Affine-Scaled Attention. Detailed training settings are described in Section \ref{sub:exp-setting}.

\subsection{Attention weights}



To empirically assess whether query-adaptive normalization yields more diverse attention behavior, we analyze the distribution of per-query cumulative attention weights. For each query token, we sum attention weights over all non-sink tokens and average across heads.

\begin{figure}[t]
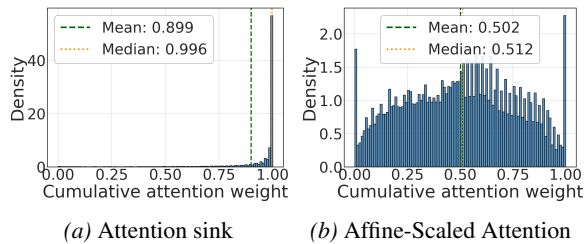

    \centering
    \begin{subfigure}[t]{0.46\linewidth}
        \centering
        \includegraphics[width=\linewidth]{fig_icml2026/attn_weight_sum_sink.pdf}
        \caption{Attention sink}
        \label{fig:attn-weight-sum-sink}
    \end{subfigure}
    \begin{subfigure}[t]{0.46\linewidth}
        \centering
        \includegraphics[width=\linewidth]{fig_icml2026/attn_weight_sum_ra.pdf}
        \caption{Affine-Scaled Attention}
        \label{fig:attn-weight-sum-ra}
    \end{subfigure}
    \caption{Per-query attention weight sum distributions for (a) Attention sink and (b) Affine-Scaled Attention. Dashed lines indicate the mean and median.} 
    \label{fig:attn-weight-hist}
\end{figure}

Figure~\ref{fig:attn-weight-hist} shows histograms of cumulative attention weights per query. By definition, standard softmax attention yields exactly one because attention weights are normalized to sum to one, which can induce unwanted concentration on first tokens. Attention sinks allow values below one by assigning probability mass to the sink, but the attainable range remains limited because the sink term is fixed and input-independent. In contrast, Affine-Scaled Attention computes a query-conditioned scale and bias and modulates the softmax outputs accordingly. This produces a substantially broader distribution, indicating greater capacity to modulate attention mass in a context-sensitive manner.


\subsection{First token analysis}

\begin{figure}[t]
  \vskip 0.2in
  \centering

  \includegraphics[width=0.45\textwidth]{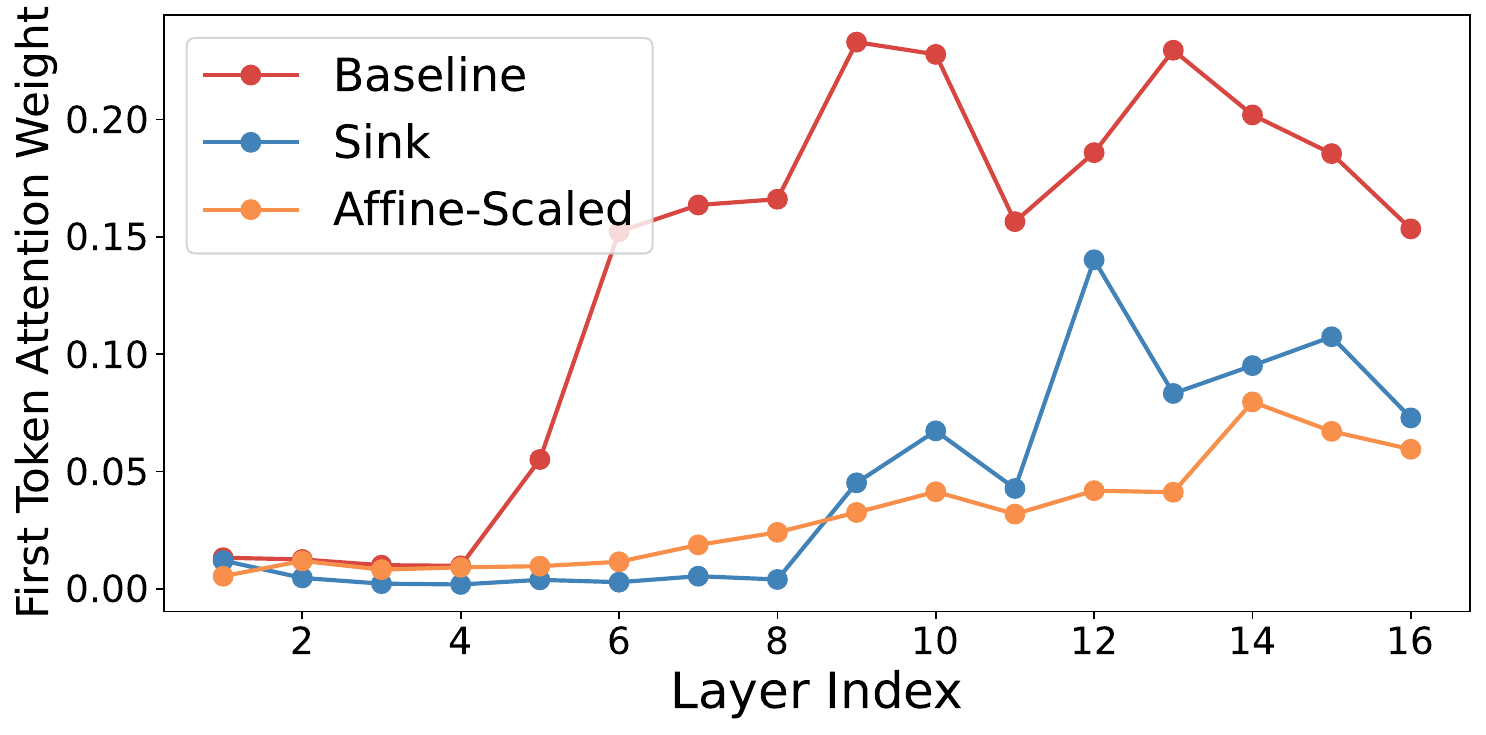}

  \caption{%
    Per-layer first-token attention weights for the 1B model. 
  }
  \label{fig:first_token_attn_graph}
\end{figure}

Greater flexibility in attention weighting can reduce the tendency to overemphasize the first token, a behavior often induced by unit-sum normalization. 
Figure~\ref{fig:first_token_attn_graph} shows the average layer-wise attention weight assigned to the first token across methods (see Appendix~\ref{app:attn_heat_map} for representative attention maps).
Standard softmax attention exhibits strong concentration on the first token. Attention sinks mitigate this effect by introducing a sink token, which reduces the attention weight on the first token, but the attention maps still show substantial concentration on the first two positions, namely the sink token and the first token. In contrast, Affine-Scaled Attention further alleviates this positional over-concentration through input-adaptive modulation. The larger reduction achieved by Affine-Scaled Attention suggests that its learned weights better reflect token-level relevance rather than position-driven concentration.

\begin{figure}[t]
  \vskip 0.2in
  \centering

  \includegraphics[width=0.48\textwidth]{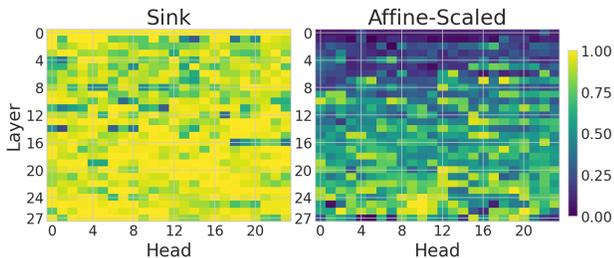}

  \caption{%
    Layer-wise and head-wise heatmaps for the 3B model showing effective attention reweighting across methods. 
  } 
  \label{fig:alpha-heatmap}
\end{figure}

\subsection{Head-wise allocation dynamics}
\label{sub:head-sparsity}

The key advantage of Affine-Scaled Attention is that it dynamically assigns head-specific scaling factors conditioned on the input, encouraging different heads to take on complementary roles rather than behaving uniformly. This added flexibility allows the attention module to respond more effectively to input characteristics at inference time, improving both information aggregation and token-to-token interactions. To examine whether this behavior aligns with our design objective, we visualize the distribution of scaling factors across layers and heads.

Figure~\ref{fig:alpha-heatmap} shows layer-wise and head-wise heatmaps of the total attention weights for Attention Sink and Affine-Scaled Attention, normalized to the range $[0,1]$. For Attention Sink, we exclude the sink value $S$ from the unit-sum normalization (i.e., we normalize over $1-S$). Attention Sink exhibits many heads saturating near the upper bound, with relatively limited variation among heads within the same layer. In contrast, Affine-Scaled Attention displays substantially richer diversity across both layers and heads, consistent with more flexible head-wise modulation. Overall, these results suggest that Affine-Scaled Attention promotes more balanced and input-adaptive utilization of attention heads compared to Attention Sink.

\begin{figure}[t]
  \vskip 0.2in
  \centering

  \includegraphics[width=0.43\textwidth]{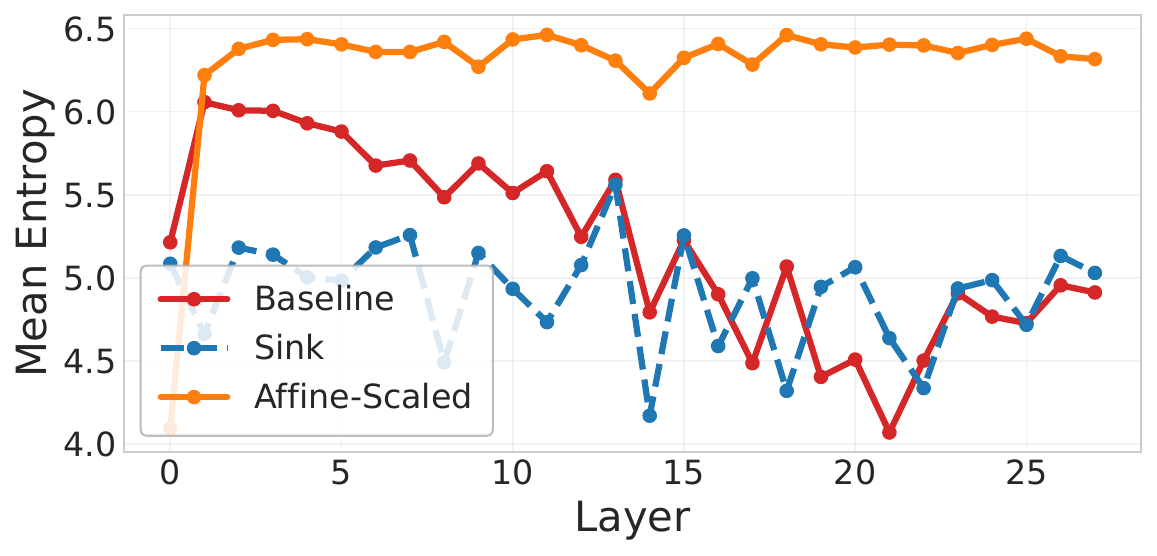}

  \caption{%
    Layer-wise mean attention entropy for the 3B model across baseline, attention sink, Affine-Scaled Attention methods. 
  } 
  \label{fig:attention-entropy}
\end{figure}

\subsection{Attention entropy}
Building on the head sparsity analysis in the previous section, we observe that relaxing the standard softmax normalization leads to more consistent utilization of attention heads.
Such additional flexibility increases attention entropy, indicating that attention mass is less concentrated on small subset of tokens or heads and is instead distributed more broadly across the sequence and across heads. 
Qualitatively, the higher entropy suggests that the model can learn more balanced and context-dependent attention patterns, rather than exhibiting position-driven concentration (e.g., over-attending to the initial token).
Figure~\ref{fig:attention-entropy} shows layer-wise attention entropy for each method. Affine-Scaled Attention achieves the highest entropy in most layers, with the largest gains appearing in deeper layers.


\begin{figure*}[t]
    \centering
        \begin{subfigure}[t]{0.3\linewidth}
        \centering
        \includegraphics[
  width=\linewidth,
]{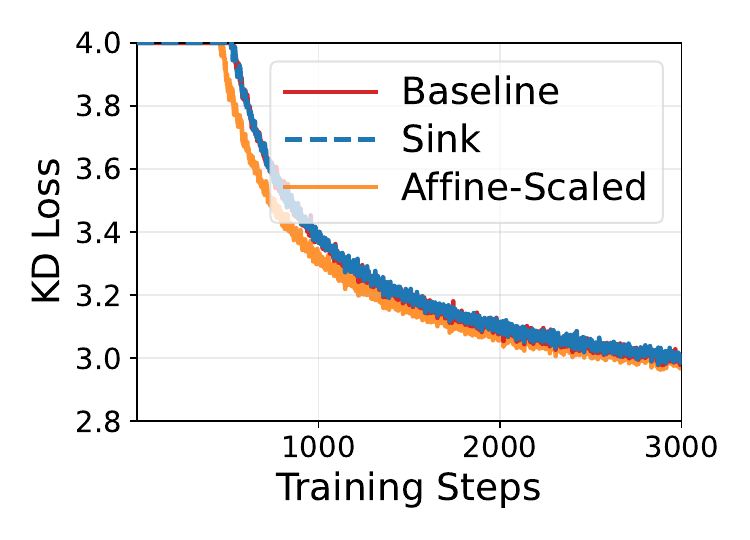}
        \caption{0.5B model training loss}
        \label{fig:kld-0.5b}
    \end{subfigure}
    \hfill
    \begin{subfigure}[t]{0.3\linewidth}
        \centering
        \includegraphics[width=\linewidth]{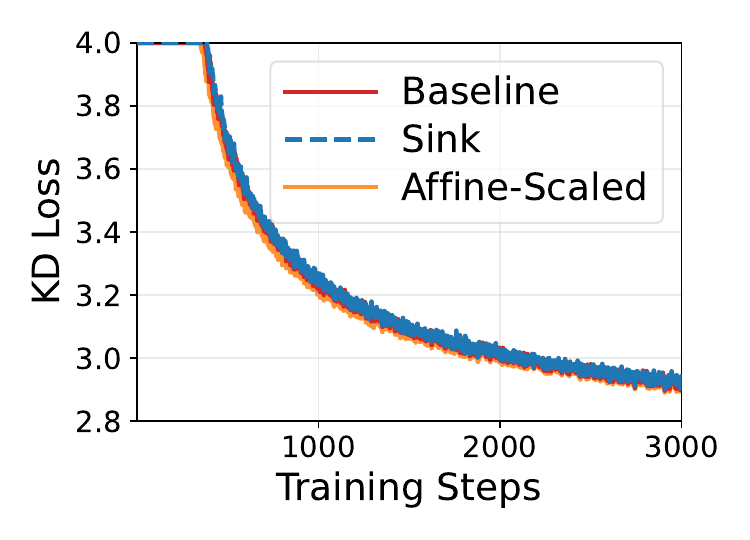}
        \caption{1B model training loss}
        \label{fig:kld-1b}
    \end{subfigure}
    \hfill
    \begin{subfigure}[t]{0.3\linewidth}
        \centering
        \includegraphics[width=\linewidth]{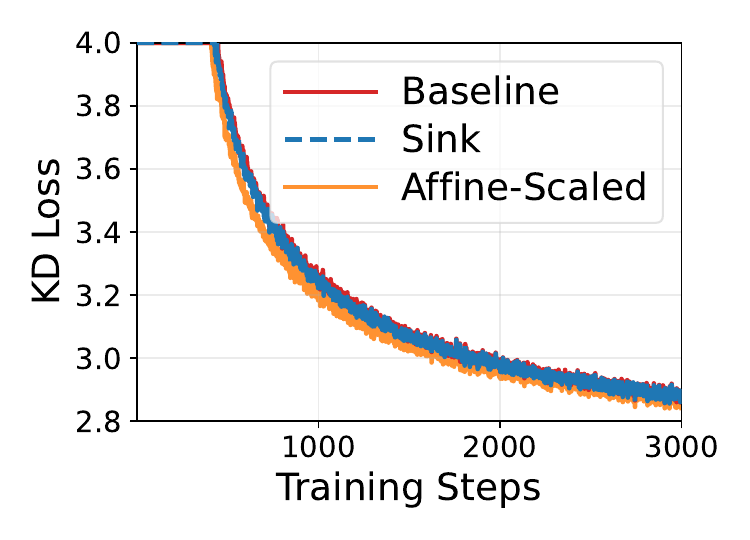}
        \caption{3B model training loss}
        \label{fig:kld-3b}
    \end{subfigure}
    \caption{KD training loss curves over training steps for (a) 0.5B, (b) 1B, and (c) 3B models across baseline, attention sink, and Affine-Scaled Attention methods, showing that Affine-Scaled Attention consistently achieves lower loss throughout training.}
    \label{fig:kld-loss-graph}
\end{figure*}

\begin{table*}[!t]
\centering
\small
\caption{Comparison of baseline, attention sink, and Affine-Scaled Attention methods for 0.5B/1B/3B models. We report the parameter overhead ($\Delta$ Params), training Loss, C4 perplexity, and downstream accuracies on common sense reasoning tasks, along with the overall average. Best accuracy within each model size is highlighted in bold.}
\resizebox{\textwidth}{!}{%
\begin{tabular}{l l r r r r r r r r r r}
\toprule
 &  &  & \multicolumn{2}{c}{} & \multicolumn{6}{c}{Accuracy (\%)} & \multicolumn{1}{c}{} \\
\cmidrule(lr){6-11}
Model & Method & $\Delta$ Params (M) & Loss & C4 PPL & Wino & PIQA & Hella & ARC-c & ARC-e & BoolQ & Avg. \\
\midrule

\multirow{3}{*}{0.5B}
 & Baseline      & 0.0 &         \textbf{2.82}  &         71.7  &         52.6  &         68.6  & \textbf{45.1} &         27.6  &         51.5  &         58.6  &         50.6  \\
 & Sink          & 0.0 & 2.83 & 81.5 & \textbf{54.9} &         68.2  &         44.3  &         26.7  &         51.9  &         58.7  &         50.8  \\
 & Affine-Scaled & 0.4 &         \textbf{2.82}  &         \textbf{70.6}  &         53.2  & \textbf{68.8} & \textbf{45.1} & \textbf{28.2} & \textbf{52.5} & \textbf{60.2} & \textbf{51.3} \\

\midrule
\multirow{3}{*}{1B}
 & Baseline      & 0.0 & 2.73 &         61.9  &         55.6  &         72.4  &         53.1  & \textbf{35.2} &         62.5  &         61.5  &         56.7  \\
 & Sink          & 0.0 & 2.73 & 68.3 &         56.6  & \textbf{72.7} &         53.2  &         34.0  &         62.0  & \textbf{61.7} &         56.7  \\
 & Affine-Scaled & 1.0 &         \textbf{2.72}  &         \textbf{58.9}  & \textbf{58.6} &         71.7  & \textbf{53.9} & \textbf{35.2} & \textbf{63.1} &         60.8  & \textbf{57.2} \\

\midrule
\multirow{3}{*}{3B}
 & Baseline      & 0.0 & 2.65 &         44.9  &         59.0  &         75.1  &         60.8  &         38.4  &         68.1  &         63.6  &         60.8  \\
 & Sink          & 0.0 & 2.65 & 48.5 & \textbf{61.2} & \textbf{75.7} &         60.9  &         40.2  &         68.1  &         63.5  &         61.6  \\
 & Affine-Scaled & 2.1 &         \textbf{2.64}  &         \textbf{40.8}  &         61.0  &         74.8  & \textbf{62.0} & \textbf{41.5} & \textbf{69.9} & \textbf{65.1} & \textbf{62.4} \\

\bottomrule
\end{tabular}%
}
\label{tab:main_results}
\end{table*}



\section{Experiments}

In this section, we provide a comprehensive comparison of several attention modulation methods based on large-scale Transformer pre-training, followed by extensive evaluation on language modeling tasks. Specifically, we evaluate a baseline model, attention sink, and Affine-Scaled Attention.




\subsection{Experimental settings}
\label{sub:exp-setting}

\paragraph{Models}
Our experiments leverage knowledge distillation as a cost-efficient and effective paradigm for pretraining language models, and are conducted using student models with 0.5B, 1B, and 3B parameters. The 0.5B student is distilled from Qwen1.5-1.8B, whereas the 1B and 3B students are distilled from LLaMA3.2-3B and LLaMA3.3-8B, respectively. All student models are trained from scratch. More detailed training settings are described in Appendix \ref{app:experimental-setting}. 

\paragraph{Hyperparameters}
We systematically sweep the learning rate from $5\times10^{-4}$ to $5\times10^{-3}$ and select $1\times10^{-3}$ for all experiments. We fix the sequence length to 2048, use a global batch size of 1024, and apply weight decay of 0.1. All models are trained on the 20B-token FineWebEdu dataset \citep{lozhkov2024fineweb-edu}. Unless otherwise specified, we set the momentum coefficient $\rho$ to 0.9 consistently.

\paragraph{Tasks}
To evaluate commonsense reasoning, we report zero-shot performance on ARC-Challenge (ARC-c) ~\cite{clark2018think}, ARC-Easy (ARC-e)~\cite{clark2018think}, HellaSwag~\cite{zellers2019hellaswag}, Phrase-Indexed Question Answering (PIQA)~\cite{bisk2020piqa}, BoolQ ~\cite{clark2019boolq} and WinoGrande~\cite{sakaguchi2021winogrande} using lm-eval-harness \citep{eval-harness} v0.4.9. We compare against the standard training setup and an attention sink baseline.




\subsection{Distillation experiment results}

Figure~\ref{fig:kld-loss-graph} compares the training loss curves under different attention mechanisms.
Across all model sizes, Affine-Scaled Attention consistently achieves lower loss throughout training.
The gap persists through convergence, suggesting improved optimization rather than a short-lived advantage in early iterations.
This trend is reflected in the final training loss reported in Table~\ref{tab:main_results}, and additional results in the Appendix~\ref{app:loss-graph} confirm that the improvement is sustained over the full training schedule.
Overall, these results indicate that Affine-Scaled Attention enables more efficient training than existing attention variants.

Table~\ref{tab:main_results} also reports zero-shot evaluation results on language modeling benchmarks.
Affine-Scaled Attention yields consistent improvements over both the baseline and the attention-sink variant across all model sizes, achieving the best average accuracy in each setting.
The gains are particularly pronounced on reasoning-oriented benchmarks at larger scales: for the 3B model, ARC-c improves from 38.4 to 41.5 and BoolQ from 63.6 to 65.1.

\subsection{Training stability}

\begin{figure}[t]
    \centering
    \begin{subfigure}[t]{0.46\linewidth}
        \centering
        \includegraphics[width=\linewidth]{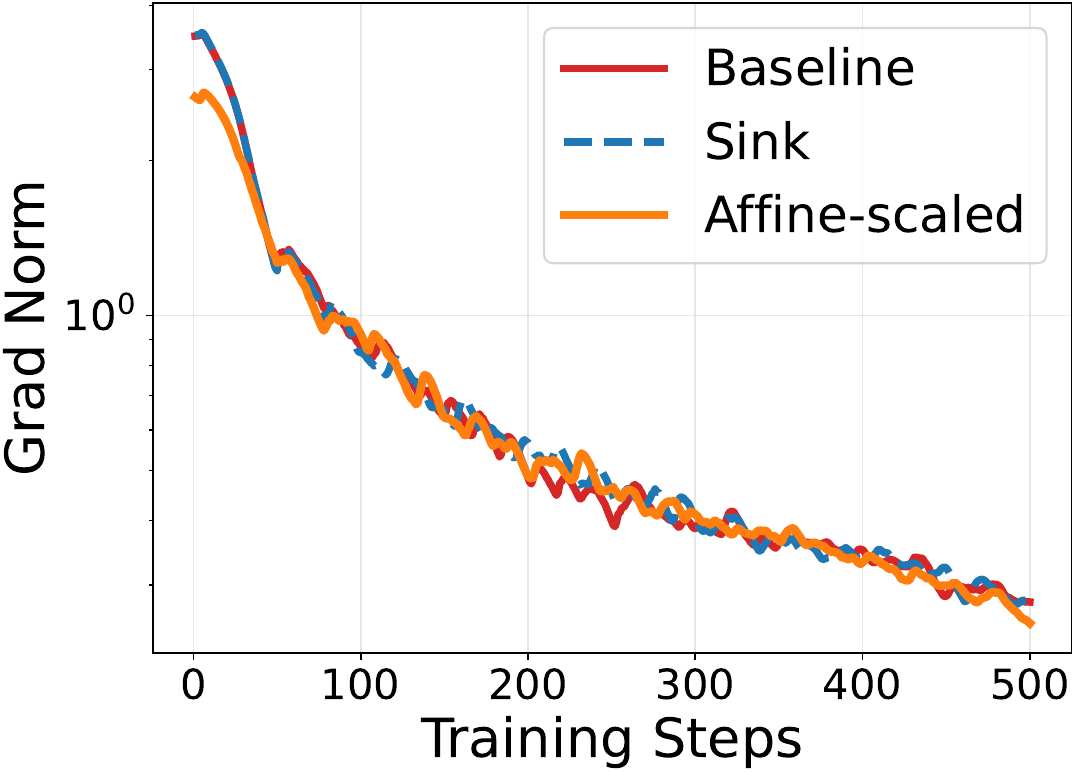}
        \caption{0.5B model gradient norm}
        \label{fig:grad-norm-0.5b}
    \end{subfigure}
    \begin{subfigure}[t]{0.46\linewidth}
        \centering
        \includegraphics[width=\linewidth]{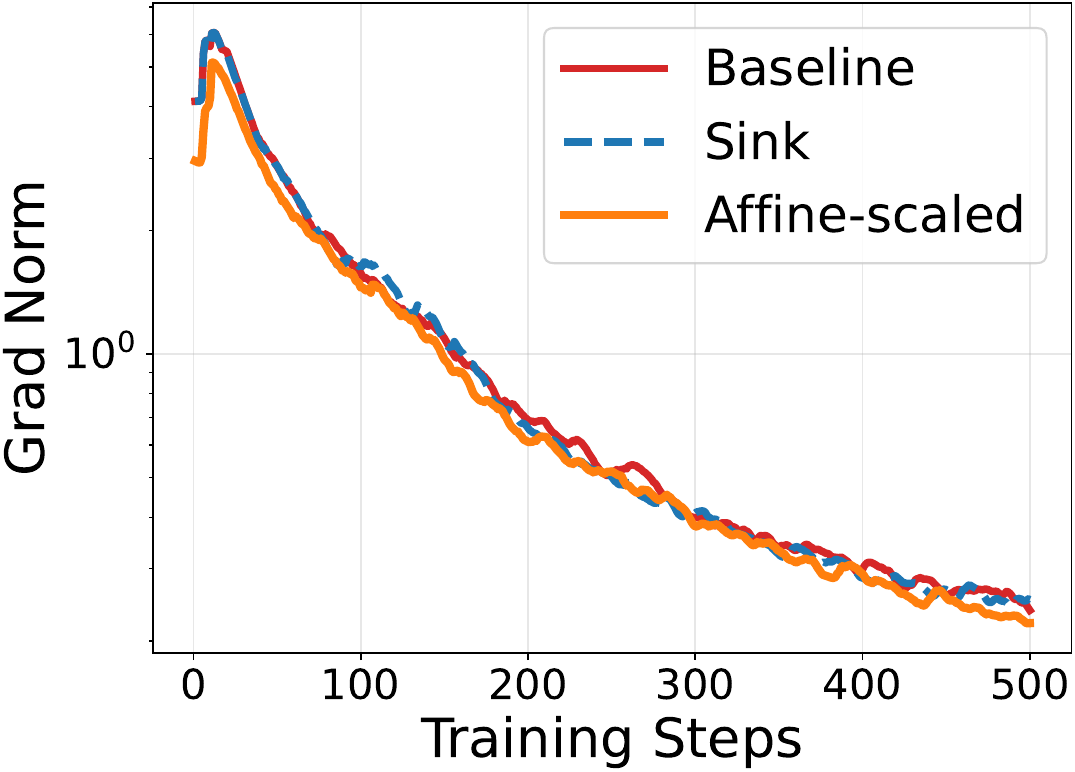}
        \caption{1B model gradient norm}
        \label{fig:grad-norm-1b}
    \end{subfigure}
    \caption{Gradient norm dynamics during training for (a) 0.5B and (b) 1B models across baseline, attention sink, and Affine-scaled Attention.}
    \label{fig:grad-norm-graph}
\end{figure}

Figure~\ref{fig:grad-norm-graph} compares the gradient norm trajectories of different attention mechanisms. Affine-Scaled Attention exhibits smoother and smaller gradient norm trajectories across various model sizes, suggesting that it contributes to more stable training dynamics. Additional analysis is provided in Appendix~\ref{app:gradient_norm}

\subsection{Affine-Scaled and Gated Attention Orthogonality}
\begin{table*}[!t]
\centering
\small
\caption{Comparison of baseline, Gated Attention, and Affine-Scaled + Gated Attention methods for 1B and 3B models. We report parameter overhead ($\Delta$ Params), training loss, C4 perplexity, and downstream accuracies on common sense reasoning tasks, along with the average. Bold indicates the best result within each model size.}
\resizebox{\textwidth}{!}{%
\begin{tabular}{l l r r r r r r r r r r}
\toprule
\multicolumn{3}{c}{} & \multicolumn{2}{c}{} & \multicolumn{7}{c}{Accuracy (\%)} \\
\cmidrule(lr){6-12}
Model & Method & $\Delta$ Params (M) & Loss & C4 PPL & Wino & PIQA & Hella & ARC-c & ARC-e & BoolQ & Avg \\
\midrule
\multirow{3}{*}{1B}
 & Baseline              &  0.0 & 2.73 & 61.9 &         55.6  & \textbf{72.4} &         53.1  &         35.2  &         62.5  &         61.5  &         56.7  \\
 & Gated                 & 67.1 &         \textbf{2.71}  &         56.6  &         57.7  &         71.9  &         53.9  &         33.9  &         62.5  &         62.8  &         57.1  \\
 & Affine-Scaled + Gated & 68.2 &         \textbf{2.71}  &         \textbf{56.2}  & \textbf{58.1} &         72.0  & \textbf{54.5} & \textbf{35.3} & \textbf{64.3} & \textbf{64.0} & \textbf{58.0} \\
\midrule
\multirow{3}{*}{3B}
 & Baseline              &   0.0 & 2.65 & 44.9 &         59.0  & \textbf{75.1} &         60.8 &          38.4  &         68.1  &         63.6  &         60.8  \\
 & Gated                 & 264.2 & 2.65 &         44.6  &         61.5  &         74.4  &         61.4 &          40.4  &         68.8  &         63.1  &         61.6  \\
 & Affine-Scaled + Gated & 266.3 &         \textbf{2.63}  &         \textbf{42.7}  & \textbf{63.4} &         74.6  & \textbf{62.7} & \textbf{41.2} & \textbf{70.0} & \textbf{66.4} & \textbf{63.1} \\
\bottomrule
\end{tabular}%
}
\label{tab:gated_results}
\end{table*}



Since Gated Attention acts as a post-attention filter that modulates propagation of the attended signal, it does not directly alter the softmax attention distribution.
As a result, Gated Attention is orthogonal to Affine-Scaled Attention, which provides explicit, fine-grained control over per-head attention weighting.
To study their interaction, we apply gated attention~\cite{qiu2025gated} after the scaled dot-product attention (SDPA) computation and evaluate models with and without Affine-Scaled Attention.

Table~\ref{tab:gated_results} summarizes results on language modeling benchmarks.
Across model sizes, adding Affine-Scaled Attention on top of Gated Attention consistently improves average accuracy over Gated Attention alone, while also yielding lower training loss and lower C4 perplexity.
These gains are observed across all benchmarks and become more pronounced at larger scales.
Overall, the results suggest that Affine-Scaled Attention provides gains largely independent of gating, and that the two mechanisms become increasingly complementary as model capacity grows.

\section{Discussion on design of activation function for scaling factor}

\begin{table}[t]
    \centering
    \small
    \caption{Average common sense reasoning performance of the 1B model comparing sigmoid and linear clipping nonlinear functions.}
    \label{tab:ablation-study}
    \begin{tabular}{c c c}
        \toprule
        Act. Function & Sigmoid & Linear clipping \\
        \midrule
        CSR Avg. & 57.6 & 58.0  \\
        \bottomrule
    \end{tabular}
\end{table}

\label{sub:activation_function}


In Section~\ref{sub:methodology}, we briefly mentioned that we apply a custom nonlinear function, but did not discuss it in detail. Here, we provide a concise explanation and an additional experiment to clarify its role. 

To prevent the scaling factors in Affine-Scaled Attention from collapsing to a constant value and to encourage more input-adaptive scaling, we introduce a custom nonlinear function for scale generation. Specifically, we apply this nonlinearity to the activation function $\phi$ in Eq.~\ref{eq:alpha}. A natural choice is the sigmoid function, which is smooth and bounded:
\begin{equation}
    \mathrm{sigmoid}(x) = \frac{1}{1 + e^{-x}},
\end{equation}
mapping an input $x$ to the range $(0,1)$.
However, the gradient of sigmoid peaks near the center and rapidly vanishes toward the extremes. 
When used for scale generation, this saturation often drives outputs toward the upper or lower bounds, yielding a highly concentrated distribution of scaling factors.

This is undesirable because the scaling factor directly controls the effective softmax temperature, and representing intermediate values is essential for fine-grained attention modulation. 
To address this, we design an activation function that retains the output range while introducing a linear region around the middle to better preserve intermediate scales.
Specifically, we propose the following linear clipping function:
\begin{equation}
\mathrm{linear\_clipping}(x)=
\begin{cases}
0, & x \le -5,\\
0.1x + 0.5, & -5 < x < 5,\\
1, & x \ge 5.
\end{cases}
\end{equation}
As illustrated in Appendix~\ref{app:ablation-study}, this piecewise-linear shape motivates the name \emph{linear clipping}. 
Compared to the sigmoid, it encourages greater variability in the learned scaling factors, enabling the model to adjust attention smoothness more effectively.

Table~\ref{tab:ablation-study} reports the average zero-shot performance on language-modeling benchmarks. The result indicate that the proposed custom linear-clipping activation function achieves higher accuracy than the sigmoid function.

\section{Conclusion}
We presented \textsc{Affine-Scaled Attention}, a simple modification that relaxes the unit-sum constraint of softmax by applying a query-dependent affine transformation to softmax-normalized weights, enabling stable control of attention magnitude.

Across large-scale experiments on multiple student model sizes, \textsc{Affine-Scaled Attention} consistently improved optimization and downstream accuracy over standard softmax and attention sink. These gains are associated with more flexible per-query attention mass, reduced first-token bias, more balanced head utilization, and higher attention entropy. In addition, Affine-Scaled Attention is complementary to Gated Attention, providing additional improvements when combined. Overall, the proposed method provides a practical way to improve training stability and performance.

\section{Limitations}
Due to limited computational resources, we focus on knowledge distillation to improve training efficiency for moderate-sized models.
We do not compare distillation against other training alternatives, such as ce-loss pretraining, extensive hyperparameter sweeps, or broader data scaling. In addition, our experiments are restricted to the moderate-size regime. We leave a thorough investigation of the proposed approach at larger model scales and with stronger teacher models as an important direction for future work.


\section*{Impact Statement}

This paper proposes Affine-Scaled Attention, a modification of the Transformer attention mechanism that improves training stability, attention flexibility, and optimization behavior in large language models. By enabling more adaptive and balanced attention allocation, this work may contribute to more efficient and robust model training, potentially lowering computational cost and improving the reliability of large-scale AI systems.

The techniques introduced here are general-purpose improvements to neural network architectures and do not introduce new application domains beyond existing large language model use cases. As such, the broader societal and ethical implications are expected to align with those already widely discussed in the context of large-scale machine learning models, including both potential benefits from improved AI capabilities and well-known risks related to misuse, bias, and deployment at scale. We do not identify any specific new ethical concerns uniquely arising from this work.

\bibliography{reference}
\bibliographystyle{icml2026}

\newpage
\appendix
\onecolumn
\section{Analysis}
\subsection{Analysis settings}
\label{app:analysis_settings}
Our analysis uses model checkpoints trained on 20B tokens. We compute the reported metrics on the C4 validation datasets with a sequence length of 1k, using 10 sampled sequences.

\subsection{Attention heatmap}
\label{app:attn_heat_map}

\begin{figure*}[h]
    \centering
        \begin{subfigure}[t]{0.3\linewidth}
        \centering
        \includegraphics[
  width=\linewidth,]{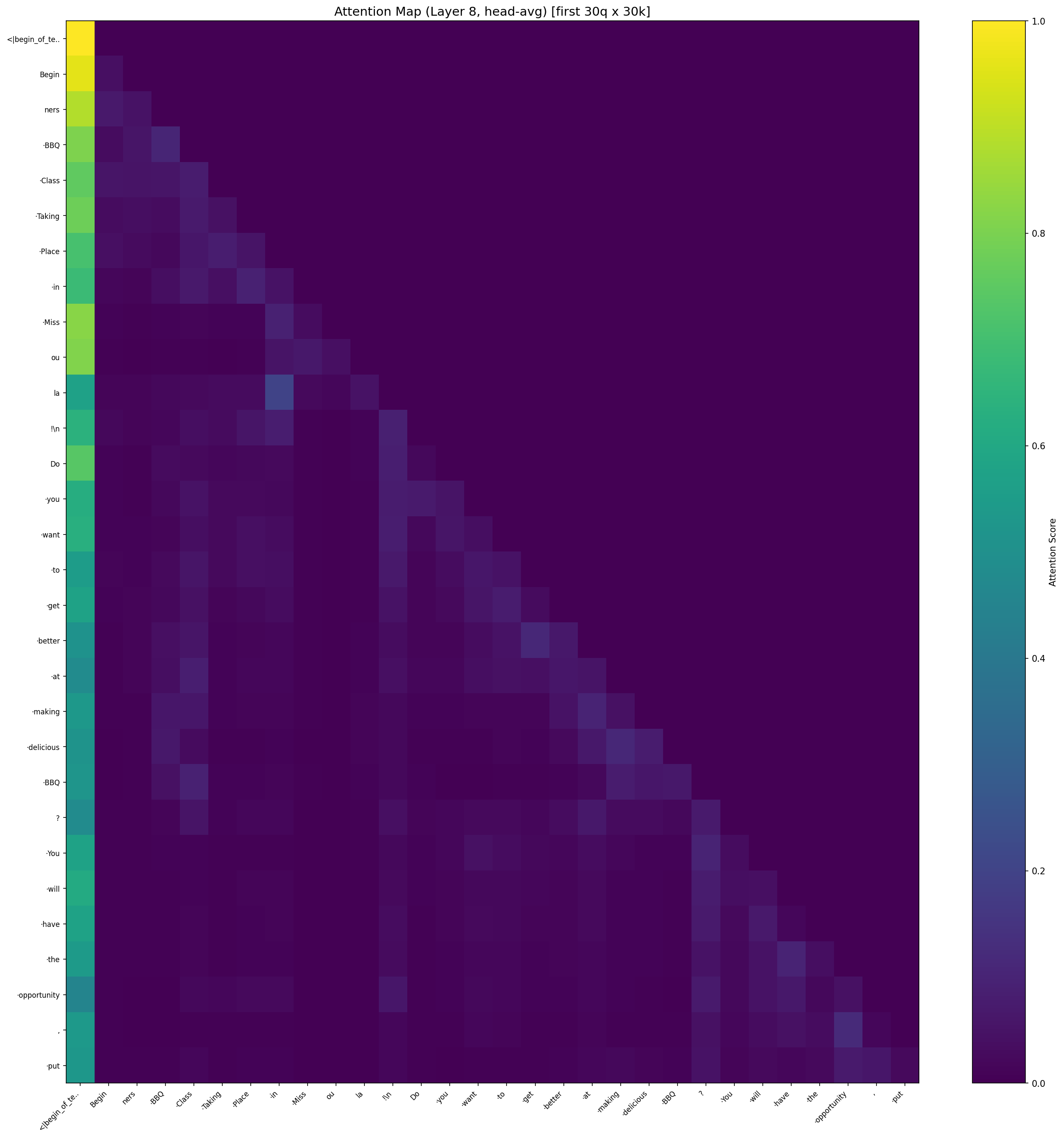}
        \caption{Baseline}
        \label{fig:base-attn-map}
    \end{subfigure}
    \hfill
    \begin{subfigure}[t]{0.3\linewidth}
        \centering
        \includegraphics[width=\linewidth]{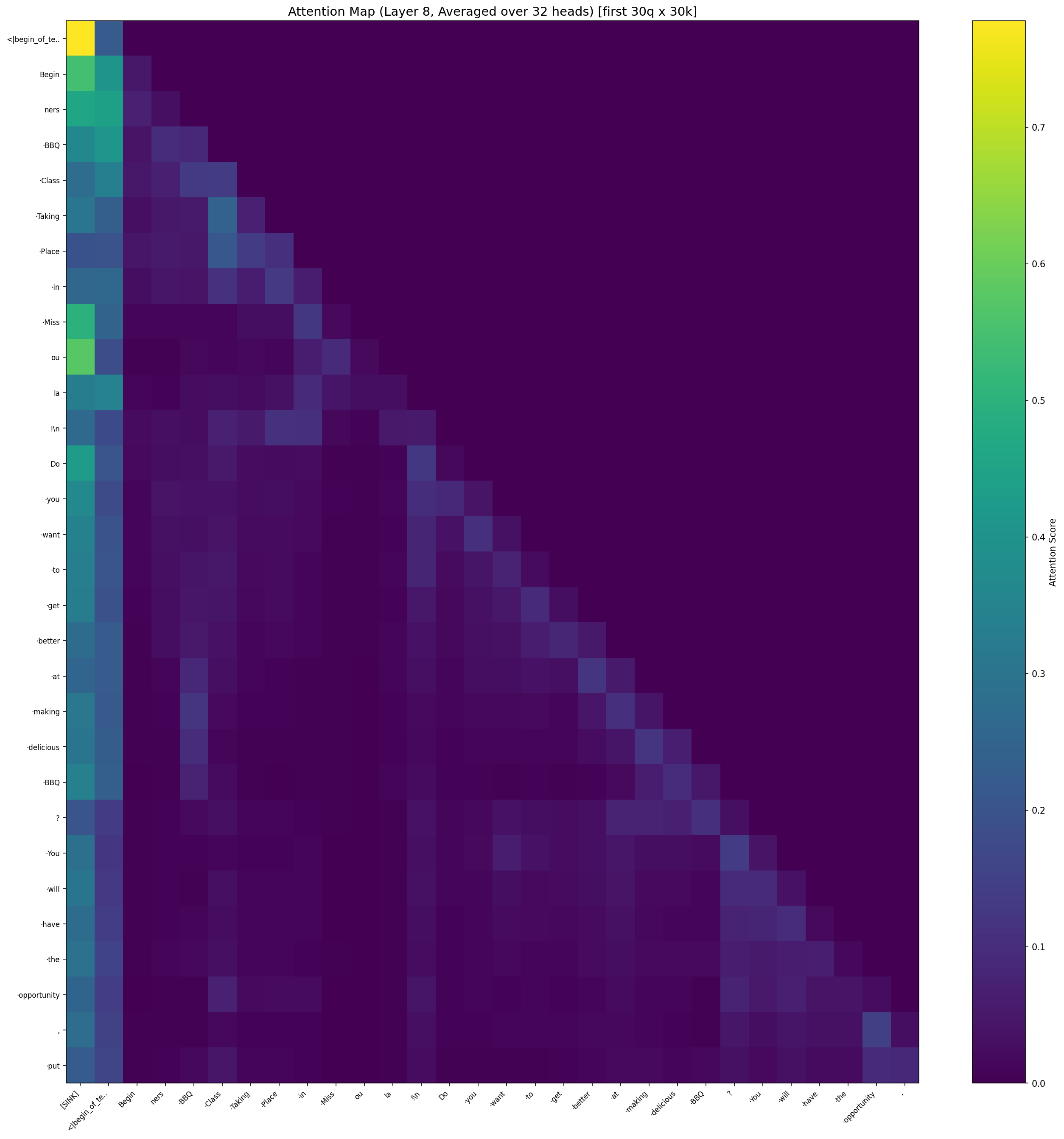}
        \caption{Attention sink}
        \label{fig:sink-attn-map}
    \end{subfigure}
    \hfill
    \begin{subfigure}[t]{0.3\linewidth}
        \centering
        \includegraphics[width=\linewidth]{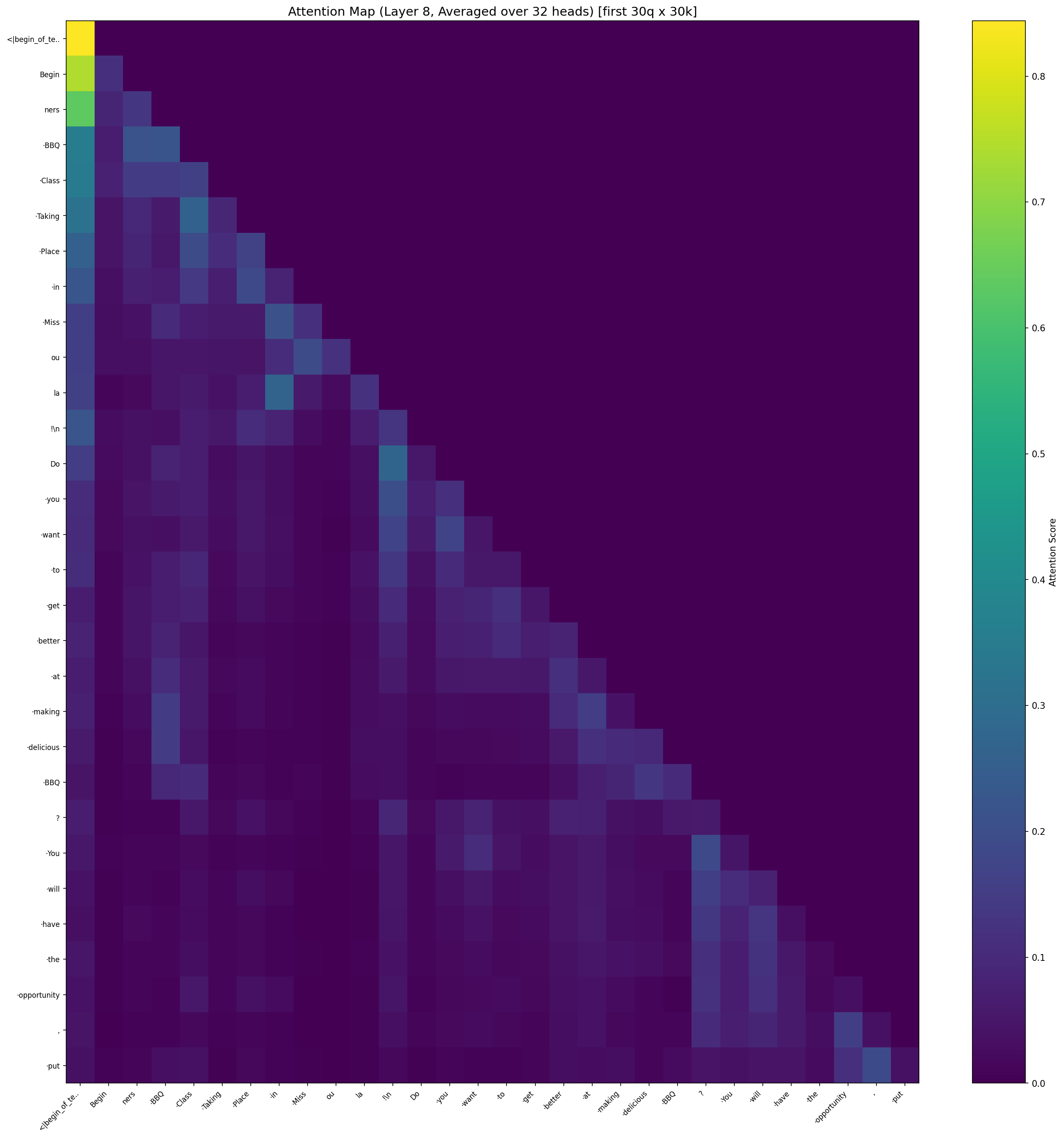}
        \caption{Affine-Scaled Attention}
        \label{fig:ra-heat-map}
    \end{subfigure}
    \caption{\textbf{Attention heatmaps for the 8th layer of the 1B model, across (a) Baseline, (b) Attention sink, and (c) Affine-Scaled Attention. }The baseline concentrates attention on the first token. Attention sink reduces the weight on the first token, but the maps still show substantial concentration on the first two positions (the sink token and the first token). In contrast, Affine-Scaled Attention mitigates this positional over-concentration via input-adaptive modulation.}
    \label{fig:attn-heat-map}
\end{figure*}

Figure~\ref{fig:attn-heat-map} visualizes the attention weight matrices from 8th layer of the 1B model, where we average the attention probabilities over heads and plot the first 30$\times$30 token block for readability. Compared to the baseline, both attention sink and Affine-Scaled Attention exhibit a more reduced allocation of attention mass to first positions (left columns), indicating a systematic shift in where attention concentrates. Notably, Affine-Scaled Attention assigns comparatively lower attention mass to the first token.

\section{Experiment}









\subsection{Experimental settings}
\label{app:experimental-setting}
We trained the model for 1 epoch of total 9000 steps with weight decay of 0.1. We used the warmup-stable-decay learning rate scheduler with 100 warmup steps. The 0.5B, 1B, and 3B student models use the same architecture as Qwen1.5-0.5B, LLaMA3.2-1B, and LLaMA3.2-3B, respectively.

\subsection{Additional loss graph}
\label{app:loss-graph}
\begin{figure*}[h]
    \centering
        \begin{subfigure}[t]{0.3\linewidth}
        \centering
        \includegraphics[
  width=\linewidth,]{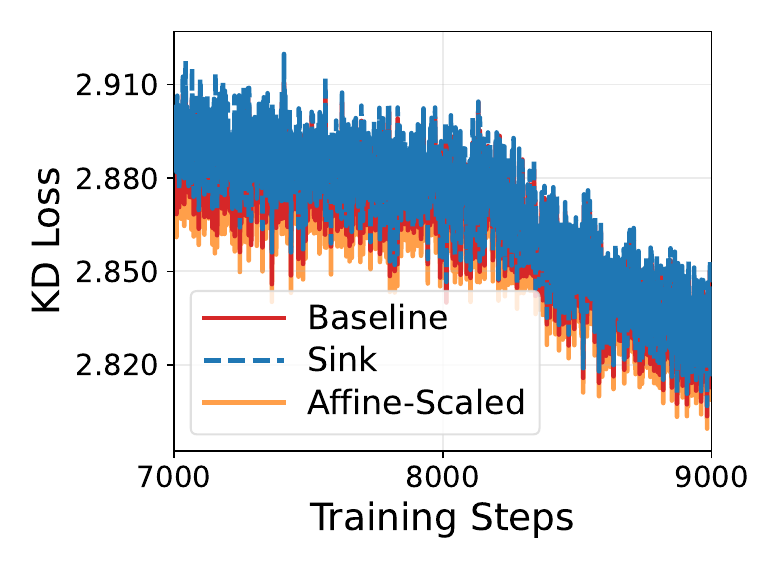}
        \caption{0.5B model}
        \label{fig:kd-loss-0.5-later}
    \end{subfigure}
    \hfill
    \begin{subfigure}[t]{0.3\linewidth}
        \centering
        \includegraphics[width=\linewidth]{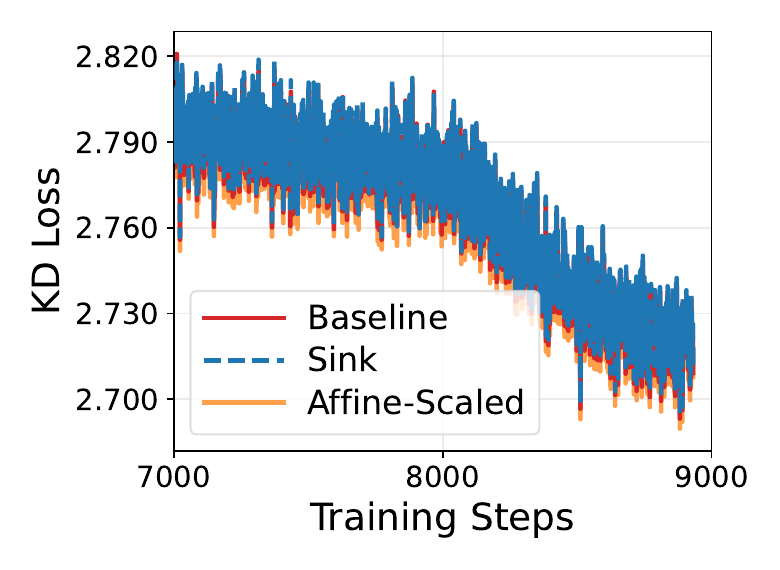}
        \caption{1B model}
        \label{fig:kd-loss-1B-later}
    \end{subfigure}
    \hfill
    \begin{subfigure}[t]{0.3\linewidth}
        \centering
        \includegraphics[width=\linewidth]{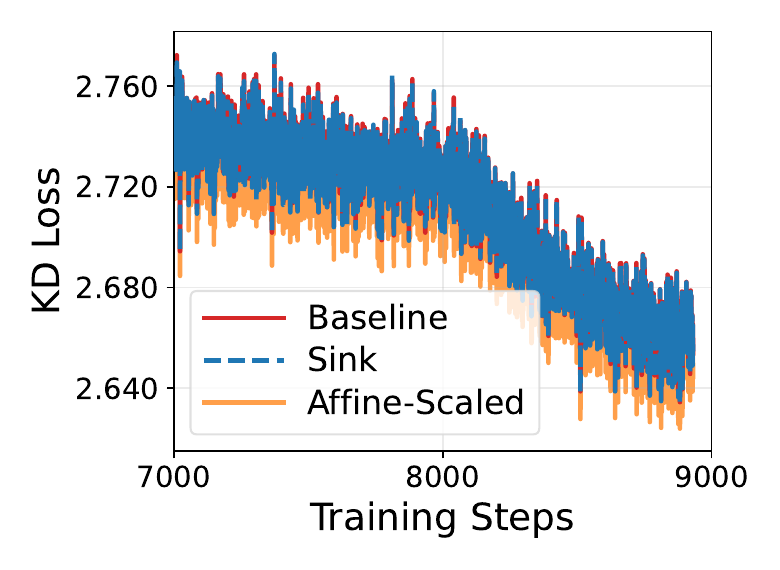}
        \caption{3B model}
        \label{fig:kd-loss-1B-later}
    \end{subfigure}
    \caption{\textbf{KD loss curves in the late training phase (steps 7k–9k) for 0.5B / 1B / 3B models. }We compare baseline, attention sink, and Affine-Scaled Attention.} 
    \label{fig:kd-loss-later}
\end{figure*}

Figure ~\ref{fig:kd-loss-later} shows knowledge distillation(KD) loss curves during the late training phase (steps 7k–9k) for the 0.5B, 1B, and 3B models. Across all model sizes, Affine-Scaled Attention consistently achieves a lower KD loss than both the baseline and attention sink methods, and this gap is maintained up to the end of training. This indicates that Affine-Scaled Attention reduces distillation error not only in the early optimization stage but also in the final stage of training.

\subsection{Gradient norm analysis}
\label{app:gradient_norm}
The results in Table~\ref{tab:grad_norm_variance} show that Affine-Scaled Attention consistently decreases gradient norm variance across different model scales. Lower variance and coefficient of variation indicate more stable gradient behavior, which may contribute to improved optimization stability in the early training phase.

\begin{table}[H]
\centering
\caption{\textbf{Gradient norm statistics during early training (steps 0--500).}
Variance and coefficient of variation (CV) are computed from raw gradient norms
without smoothing.}
\label{tab:grad_norm_variance}
\begin{tabular}{llcccc}
\toprule
Model & Method & Mean & Std & Variance & CV \\
\midrule
\multirow{3}{*}{0.5B}
 & Baseline        &         0.563  &         0.507 &          0.257  &         0.90  \\
 & Sink            & \textbf{0.551} &         0.509 &          0.259  &         0.92  \\
 & Affine-Scaled   &         0.561  & \textbf{0.451} & \textbf{0.203} & \textbf{0.80} \\
\midrule
\multirow{3}{*}{1B}
 & Baseline        &         0.907  &         1.983  &         3.934  &         2.19  \\
 & Sink            &         0.902  &         1.988  &         3.951  &         2.20  \\
 & Affine-Scaled   & \textbf{0.842} & \textbf{1.702} & \textbf{2.897} & \textbf{2.02} \\
\bottomrule
\end{tabular}
\end{table}

To further assess optimization stability beyond gradient variance, we measure the frequency of extreme gradient norm spikes during the early training phase. As shown in Table~\ref{tab:grad_spike_count}, Affine-Scaled
Attention consistently reduces the number of highly extreme gradient events,
particularly under the stricter threshold ($k=9$), across both model scales.
This suggests that Affine-scaled Attention leads to smoother gradient dynamics
and improved training stability by mitigating abrupt optimization fluctuations.

\begin{table}[H]
\centering
\caption{\textbf{Extreme gradient norm spike counts during early training (steps 0--500).}
A spike is defined as a training step where the gradient norm exceeds a robust
run-specific threshold, $\tau = \mathrm{median}(g) + k \cdot \mathrm{MAD}(g)$,
with $k \in \{6, 9\}$. MAD denotes the median absolute deviation, a robust measure of variability
that is less sensitive to outliers than standard deviation. Lower spike counts indicate fewer extreme gradient events
and smoother optimization dynamics.}
\label{tab:grad_spike_count}
\begin{tabular}{llcc}
\toprule
Model & Method & \#Spikes ($k=6$) & \#Spikes ($k=9$) \\
\midrule
\multirow{3}{*}{Qwen-0.5B}
 & Baseline        & 40 & 30 \\
 & Sink            & 44 & 31 \\
 & Affine-Scaled   & \textbf{42} & \textbf{22} \\
\midrule
\multirow{3}{*}{LLaMA-1B}
 & Baseline        & \textbf{53} & 28 \\
 & Sink            & 67 & 35 \\
 & Affine-Scaled   & 54 & \textbf{27} \\
\bottomrule
\end{tabular}
\end{table}
\newpage
\subsection{Discussion on design of activation function for scaling factor}
\label{app:ablation-study}
We define our custom activation function, \emph{linear clipping}, as follows:

\begin{figure}[h]
  \centering
  \includegraphics[width=0.5\linewidth]{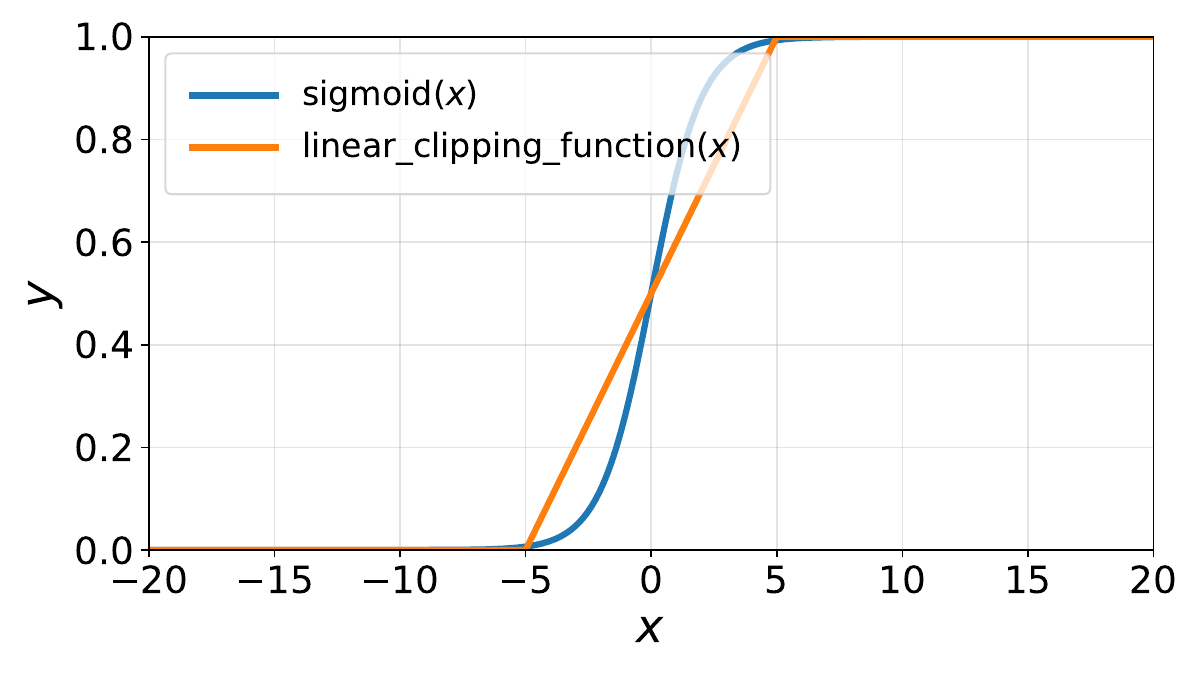}
  \caption{Comparison of a sigmoid and a custom clipping-linear function.}
  \label{fig:sigmoid_vs_linear_clipping}
\end{figure}


\end{document}